\documentclass[sigconf]{acmart}

\AtBeginDocument{%
  \providecommand\BibTeX{{%
    \normalfont B\kern-0.5em{\scshape i\kern-0.25em b}\kern-0.8em\TeX}}}

\usepackage{adjustbox}
\usepackage{multirow}
\usepackage{multicol}
\usepackage{amsmath,amsfonts}
\usepackage{algorithmic}
\usepackage{graphicx}
\usepackage{textcomp}
\usepackage{xcolor}
\usepackage{longtable}
\usepackage{multirow}
\usepackage{booktabs}
\usepackage{hyperref}
\usepackage{indentfirst}
\usepackage{lineno}
\usepackage{float}
\usepackage{cancel}
\usepackage{adjustbox}
\usepackage{tabularx}
\usepackage{lipsum}
\usepackage{xcolor}  
\usepackage{multirow}
\usepackage[normalem]{ulem}
\usepackage{caption}
\begin{document}
\title{VMambaCC: A Visual State Space Model for Crowd Counting}


\author{Hao-Yuan Ma}
\authornotemark[1]
\email{20224227066@stu.suda.edu.cn}
\affiliation{%
  \institution{School of Computer Science and Technology, Soochow University}
  \streetaddress{No. 333, Ganjungdong Road}
  \city{Suzhou}
  \state{Jiangsu}
  \country{china}
  \postcode{215000}
}

\author{Li Zhang}
\authornotemark[2]
\email{zhangliml@suda.edu.cn}
\affiliation{%
  \institution{School of Computer Science and Technology, Soochow University}
  \streetaddress{No. 333, Ganjingdong Road}
  \city{Suzhou}
  \state{Jiangsu}
  \country{china}
  \postcode{215000}
}

\author{Shuai Shi}
\authornotemark[1]
\email{20235227140@stu.suda.edu.cn}
\affiliation{%
  \institution{School of Computer Science and Technology, Soochow University}
  \streetaddress{No. 333, Ganjungdong Road}
  \city{Suzhou}
  \state{Jiangsu}
  \country{china}
  \postcode{215000}
}


\begin{abstract}
	
As a deep learning model, Visual Mamba (VMamba) has a low computational complexity and a global receptive field, which has been successful applied to image classification and detection. To extend its applications, we apply VMamba to crowd counting and propose a novel VMambaCC (VMamba Crowd Counting) model. Naturally, VMambaCC inherits the merits of VMamba, or global modeling for images and low computational cost. 
Additionally, we design a Multi-head High-level Feature (MHF) attention mechanism for VMambaCC. MHF is a new attention mechanism that leverages high-level semantic features to augment low-level semantic features, thereby enhancing spatial feature representation with greater precision. Building upon MHF, we further present a High-level Semantic Supervised Feature Pyramid Network (HS2PFN) that progressively integrates and enhances high-level semantic information with low-level semantic information.
Extensive experimental results on five public datasets validate the efficacy of our approach. For example, our method achieves a mean absolute error of 51.87 and a mean squared error of 81.3 on the ShangHaiTech\_PartA dataset. Our code is coming soon.
\end{abstract}



\maketitle

\section{Introduction}
Crowd counting refers to the task of quantifying the number of individuals in images or videos using computer vision techniques.
This task is of significant importance to predicting crowd trends and optimizing resource allocation, and can provide crucial information for many practical applications, such as urban planning \cite{P2PNet}, security surveillance \cite{GauNet}, and retail analysis \cite{FGENet}.

For crowd counting tasks, there has been challenges including high density, occlusion, and small targets. To meet these challenges, current crowd counting models are mainly constructed based on two architectures: Convolutional Neural Networks (CNNs) \cite{MCNN,GauNet,FGENet,McML,CSRNet} and Transformer \cite{CLTR,LoViT}. 

CNN is a kind of deep learning model specifically designed for processing data like images. 
There are many classic CNN models for crowd counting, such as 
Multi-Column CNN (MCNN) \cite{MCNN}, Congested Scene Recognition Network (CSRNet) \cite{CSRNet}, Context-Aware Feature Network (CANNet) \cite{CANNet}, Attention Scaling NetWork (ASNet) \cite{ASNet}, Shallow Feature Based Dense Attention Network  (SDANet) \cite{SDANet}, and Fine-Grained Extraction Network (FGENet) \cite{FGENet}. 
These models leverage CNN to extract features and fuse multi-scale features for counting individuals. However, these methods are sensitive to changes in position and scale. In addition, as Kolesnikov et al. \cite{Vit} point out that CNN-based models struggle to obtain a global receptive field; thus, these models are restricted when dealing with multi-scale individuals. 

Transformer is a kind of deep learning model originally designed for processing time sequence data. 
By dividing spatial positions and image patches into sequences, Kolesnikov et al. \cite{Vit} proposed a Visual Transformer (ViT) based on Transformer to capture relationships between pixels in images. 
Transformer-based models mainly use ViT to extract features from images. Typical models are Crowd Localization TRansformer (CLTR) \cite{CLTR} and Local features by Vision Transformer for Crowd counting (LoViTCrowd) \cite{LoViT}. These methods possess a global receptive field.
However, the computational complexity of Transformer-based models exponentially increases as the size of the input images, leading to reduced efficiency in tasks with large computational costs \cite{Mamba}.


Recently, Gu and Dao \cite{Mamba} proposed a Mamba model, which is a state space model, for natural language processing.  
Mamba enables each element in a sequence to interact with any previously scanned element, thereby reducing the quadratic complexity of the attention mechanism in Transformer to the linear complexity.
Inspired by the Mamba, Liu et al. \cite{Vmamba} proposed a Visual Mamba (VMamba) model. VMamba designs a Cross-Scan Module (CSM) that adopts a four-way scanning strategy, starting from the four corners of the image. In doing so, VMamba can have a global attention and maintain the linear computational complexity.

\begin{figure}[t]
	\centering
	\includegraphics[width=0.4\textwidth]{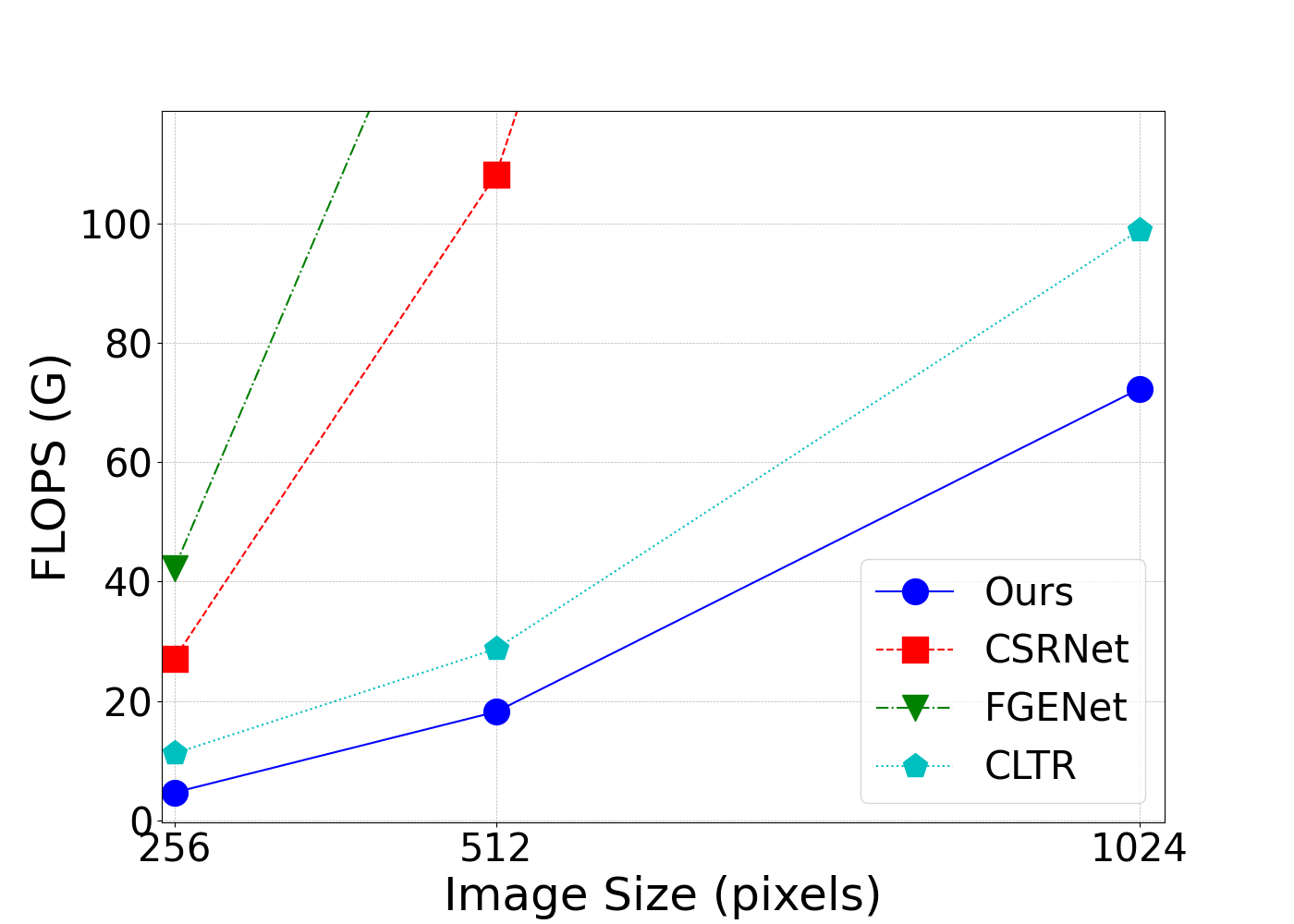}\hfill
	\caption{Comparison curves of flops vs. image size obtained by four methods. }
	\label{fig:0}
\end{figure}

Building upon VMamba, we propose a versatile VMambaCC model that is capable of crowd counting and crowd localization. VMambaCC not only captures global information but also tackles the multi-scale variations caused by the perspective effect. Our network stands out from popular Transformer-based or CNN architectures by providing linear scaling of feature sizes. 
Fig. \ref{fig:0} plots the curves of computational complexity (flops) vs. image size obtained by our method and other three  mainstream methods. These curves imply that our method is capable of handling the processing more efficiently, with the computational complexity consistently maintained at a lower level.
Our main contributions are as follows:
\begin{itemize}
\item A novel VMambaCC model is proposed, which is the first visual state space model for crowd counting that is a high-density task. VMambaCC has not only a global receptive field but also a fast  learning and inference speed.

\item VMambaCC contains a Multi-head High-level Feature (MHF) enhancement mechanism that is newly designed for enhancing low-level semantic information by reinforcing high-level semantic features.

\item On the basis of the MHF attention mechanism, this study presents a feature fusion network, or the High-level Semantic Supervised Feature Pyramid Network (HS2FPN). This network gradually integrates information from different scales and partially preserves high-level semantic features.
\end{itemize}

The subsequent sections of this paper are structured as follows.
In Section 2, we provide a concise review of the relevant literature pertaining to crowd counting tasks. Section 3 delves into a detailed exposition of the VMambaCC model, encompassing its architecture, design philosophy, and the innovative components including the MHF attention mechanism and the HS2FPN feature fusion network. Section 4 presents the results of our counting, localization, and ablation studies, offering a comprehensive evaluation and analysis of our model. Finally, Section 5 summarizes this paper.

\section{Related work}
Crowd counting has attracted widespread attention in the field of computer vision. For decades, methods for crowd counting  include detection-based \cite{jiance1,jiance2,jiance3}, density map-based \cite{MCNN,CSRNet,CANNet,GauNet,GLoss,LoViT,MAN,McML}, and point-based approaches \cite{FGENet,P2PNet,CLTR}.

Detection-based methods transform the crowd counting task into an object detection problem by detecting individuals in the image for counting. These methods typically utilize object detection algorithms such as faster region-CNN \cite{FasterRCNN} and YOLO \cite{yolox} to detect individuals in an image and count them based on the detected instances. Although detection-based methods can provide more accurate localization and counting of individuals, their performance may be affected when dealing with high-density and heavily occluded scenarios.

On the basis of the principle of density estimation, density map-based methods generate a grayscale image, where each pixel represents the estimated density of individuals at corresponding locations. 
Typically, density estimation is performed using some schemes, such as Gaussian kernel functions \cite{gsk}, where each annotated point in the original image is convolved with a Gaussian kernel to obtain the density estimation for each pixel and form a density map. 
Popular density map-based methods include MCNN \cite{MCNN}, CSRNet \cite{CSRNet}, and others. However, density maps, as an estimate of the overall distribution of the crowd, lose the individual location information, leading to limitations in individual tracking and recognition. In addition, the superposition of Gaussian kernels can introduce additional noise to the data \cite{GauNet}.

Point-based methods employ an effective approach to crowd counting by representing individuals as points and estimating the total number of individuals in the image \cite{P2PNet}. Compared to other methods, the point-based framework offer better robustness and computational efficiency as it only requires detecting the positions of individuals rather than complete body detection boxes, reducing computational complexity. This kind of approaches are commonly used in crowded scenes such as stations and squares for real-time monitoring and analysis of crowd behavior, with broad application prospects. Some classic point methods include P2PNet \cite{P2PNet}, FGENet \cite{FGENet}, and CLTR \cite{CLTR}. So does our model.

\section{Proposed model}
This section first introduces the framework of VMambaCC, then  discusses the newly proposed the MHF enhancement mechanism, and finally describes the fusion network, HS2FPN.

First of all, we clarify some notations. The set of annotated points in an image is denoted as $S_a=\{(x_i,y_i)|i=1,\cdots,N\}$, where $N$ is the number of points, and $x_i$ and $y_i$ represent the horizontal and vertical coordinates of the $i$-th individual in the image, respectively. Let $S_p=\{(\hat{x}_j, \hat{y}_j, \hat{c}_j)| j =1,\cdots, M\}$ be the set of predicted points, where $\hat{x}_j$ and $\hat{y}_j$ represent the predicted horizontal and vertical coordinates of the $j$-th individual in the image, respectively,  $\hat{c}_j$ is the confidence score that the $j$-th predicted point is an individual, and $M$ is the number of predicted points.

\begin{figure*}[t]
	\centering
	\includegraphics[width=0.8\textwidth]{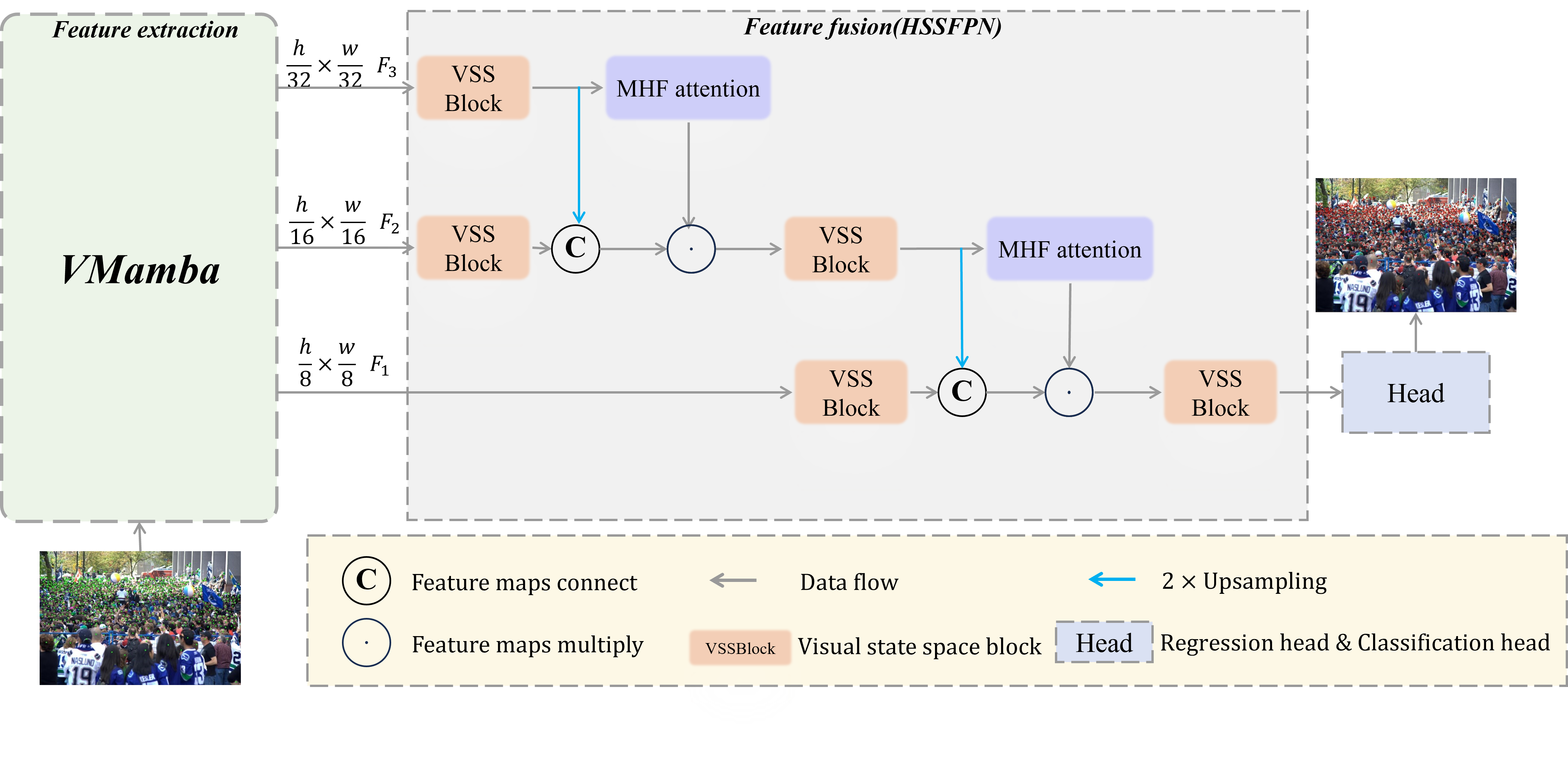}\hfill
	\caption{Architecture of VMambaCC model.}
	\label{fig1:model}
\end{figure*}
\subsection{Framework of VMambaCC} \label{seq:3_1}
VMambaCC is the first Mamba-based model specifically tailored for crowd counting tasks. Because VMambaCC has the advantages of VMamba, it addresses the issue of Transformer's quadratic growth in computational cost with image block size and retains the global receptive field. 

As shown in Fig. \ref{fig1:model}, VMambaCC consists of three stages: feature extraction, feature fusion, and prediction. The input of this model is an image, and the output is a set of predicted points with confidence scores. Therefore, VMambaCC can be taken as a kind of point-based counting model. 
For a given crowd image $\mathbf{X}$, VMamba is first used to extract its feature maps with three scales, denoted as $\mathbf{F}_s$, $s=1,2,3$.
Subsequently, these feature maps pass through the feature fusion stage. By progressively fusing features at three scales, HS2FPN preserves high-level feature supervision and captures individual information at different resolutions. 
Finally, our model outputs both positions and confidence scores of individuals in the image $\mathbf{X}$ after passing through the prediction stage.

For training VMambaCC, we utilize a Three-Task Combination (TTC) loss function that involves matching annotated points with predicted points. We treat this as a bipartite graph matching problem and use the Hungarian algorithm to match points in $S_a$ with those in $S_p$. Without loss of generality, the first $N$ predicted points in $S_p$ are matched successfully with annotated points in $S_a$ in order.

In VMambaCC, we design a novel MHF attention mechanism and then propose a fusion network (NS2PFN) based on this new attention mechanism. 
By effectively combining the semantic information from high-level features with the detailed information from low-level features, MHF can help our model improve its performance when handling multi-scale data.
In the following, we discuss them further.  

\begin{figure}[h]
    \centering
    \includegraphics[width=0.45\textwidth]{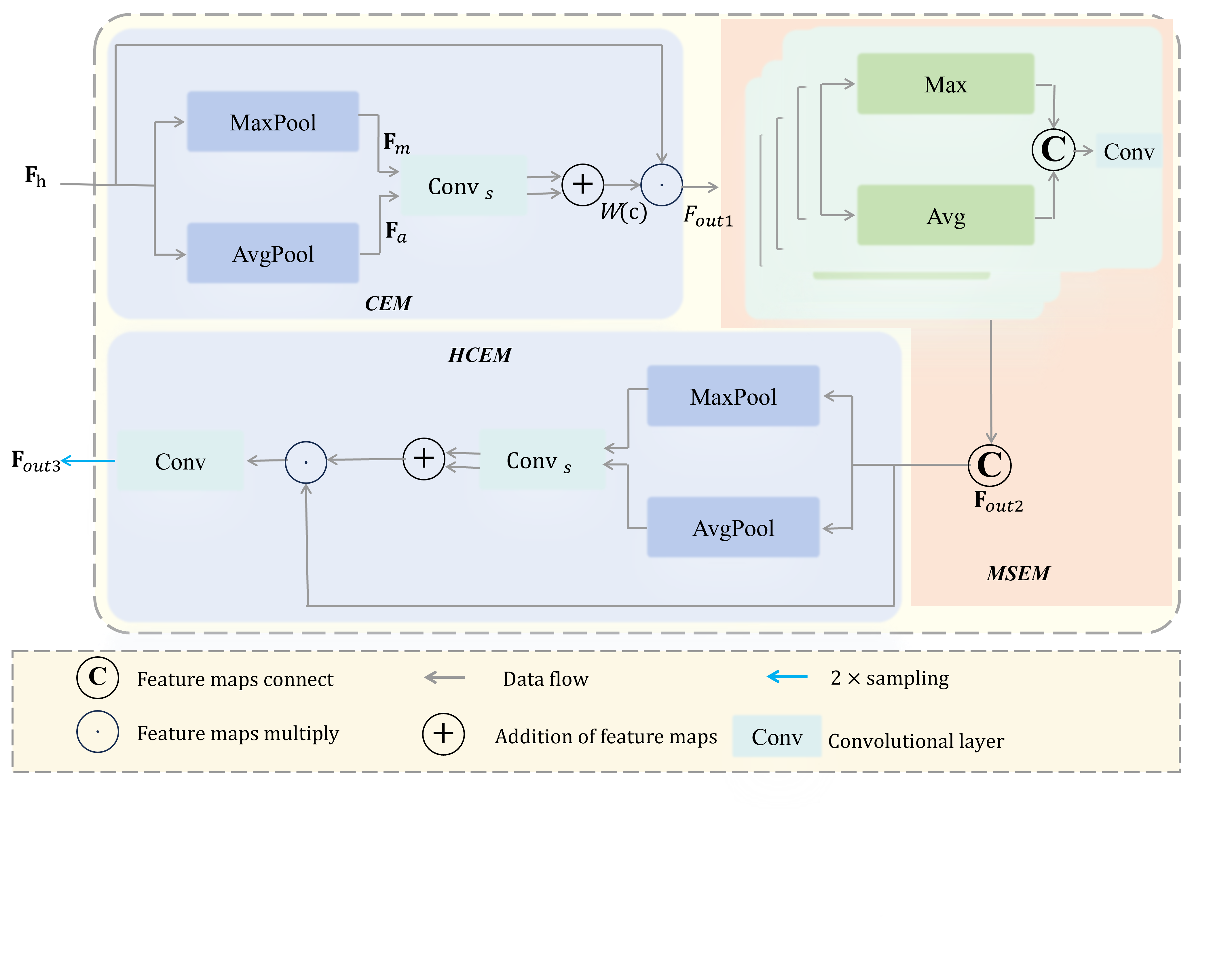}\hfill
    \caption{Structure of MHF attention mechanism.}
    \label{fig2:Attention}
\end{figure}

\subsection{MHF attention mechanism} \label{seq:3_2}

Fig. \ref{fig2:Attention} shows the structure of MHF attention. 
Roughly, the MHF attention can be divided into three parts: the Channel Enhancement Module (CEM), the Multi-head Spatial Enhancement Module (MSEM), and the High-level CEM (HCEM).

\subsubsection{CEM} The purpose of CEM is to assign attention weights to channels, allowing the model to better select and emphasize important features. In this module, the high-level feature maps are passed through max-pooling and average-pooling layers. 

Let $\mathbf{F}_h$, $\mathbf{F}_m$ and $\mathbf{F}_a$ be the high-level, max-pooled and average-pooled features, respectively. That is, 
\begin{equation}\label{eq:01}
	\mathbf{F}_{m}=MaxPool(\mathbf{F}_h)
\end{equation}
and 
\begin{equation}\label{eq:02}
	\mathbf{F}_a=AvgPool(\mathbf{F}_h)
\end{equation}
where $MaxPool(\cdot)$ and $AvgPool(\cdot)$ are the operations of max-pooling and average-pooling, respectively.

Let $\mathbf{F}_{out_1}$ be the output of CEM. Then, $\mathbf{F}_{out_1}$ can be obtained by 
\begin{equation}\label{eq:03}
	\mathbf{F}_{out_1}(c) = W(c) \mathbf{F}_h(c) 
\end{equation}
where 
$\mathbf{F}_{out_1}(c)$ and $\mathbf{F}_h(c)$ are the feature maps in the $c$-th channel of $\mathbf{F}_{out_1}$ and $\mathbf{F}_{h}$, respectively, $\mathbf{W}=[W(1),\cdots, W(c)]^T$ is the channel weight vector that is defined as 
\begin{equation}\label{eq:04}
\mathbf{W} = \text{Sigmoid}\left(Conv_{s}(\mathbf{F}_{m})+Conv_{s}(\mathbf{F}_{a})\right)
\end{equation}
$Conv_{s}(\cdot)$ represents a sequence of convolutional layers with shared weights, and $Sigmoid(\cdot)$ denotes the Sigmoid activation function.

\sloppy
\subsubsection{MSEM} The role of MSEM is to make the model focus on useful information, which could improve the robustness of the model. If we extract only the maximum and average values in feature maps among all channels, then it is highly possible to lose lots of grained information. Therefore, we divide feature maps into multiple groups and design multi-head spatial features to focus on grained information from different groups.
After obtain the output of CEM, we divide $\mathbf{F}_{out_1}$ into 4 groups according to channels. That is,
\begin{equation}\label{eq:05}
\mathbf{F}_{g_i} = Group(\mathbf{F}_{out_1}),i=1,2,3,4
\end{equation}
where $Group(\cdot)$ is the operation of dividing feature maps into groups. 

Then, the output of MSEM can be expressed by
\begin{equation}\label{eq:06}
\mathbf{F}_{out_2} = Concat\left(\mathbf{G}_1,\cdots,\mathbf{G}_4\right)
\end{equation}
where $\mathbf{G}_i=Conv(Max(\mathbf{F}_{g_i}),Avg(\mathbf{F}_{g_i}))$, $Concat(\cdot)$ is the feature map concatenation, $Conv(\cdot)$ filters the extracted feature maps, $Max(\cdot)$ and $Avg(\cdot)$ are functions of taking the maximum and average values among the channels, respectively.

\subsubsection{HCEM} The main purpose of this module is to enhance important channels and upsample feature maps. 
After MSEM, there may be some redundant information. In order to filter out the redundant information, we need to enhance $\mathbf{F}_{out_2}$ in terms of channels, then fuse the feature maps and multiply them onto the low-level features. 

\begin{eqnarray}\label{eq:07}
\mathbf{F}_{out3} &=& Upsample(Conv(Conv_{s}(MaxPool(\mathbf{F}_{out_2}))\\
& &+Conv_{s}(AvgPool(\mathbf{F}_{out_2}))))
\end{eqnarray}
where $Upsample(\cdot)$ stands for upsampling.

\begin{table*}[htbp]
    \caption{Statistics of crowd counting datasets.}\label{tab:data}
    \centering
        \begin{tabular}{lcccccc}
            \hline
            Dataset & \#Number of images  & \#Training/Validation/Test & Total & Min & Average & Max \\ \hline
            SHT\_A & 482  & 300/0/182 & 241,677 & 33 & 501.4 & 3139 \\ 
            SHT\_B & 716  & 400/0/316 & 88,488 & 9 & 123.6 & 578 \\ 
            UCF\_QNRF & 1,535  & 1201/0/334 & 1,251,642 & 49 & 815 & 12,865 \\ 
            JHU-CROWD & 4250  & 3888/0/1062 & 1,114,785 & - & 262 & 7286 \\ 
            UCF\_CC\_50 & 50  & - & 63,974 & 94 & 1,280 & 4,543 \\ \hline
        \end{tabular}
\end{table*}

\subsection{HS2FPN} \label{seq:3_3}
We make a thorough analysis on existing crowd counting models and find that existing feature fuse networks have a limitation when dealing with high-density and complex scenes. 
The main constraints lie in the insufficient global perception capability of these networks, which fails to effectively capture the overall information in images, and the inadequate extraction and representation of key features in scenes, especially when dealing with occlusion and varying scales of crowds. 
Therefore, HS2FPN aims to address these issues, thereby improving the accuracy and robustness of crowd counting.

HS2FPN mainly consists of multiple Visual State Space (VSS) and MHF attention blocks. 
VSS, a module from VMamba \cite{Vmamba}, can effectively enhance the model's perception of global information in images through a four-way scanning strategy, enabling a more comprehensive understanding of scenes and improving the capability to capture distribution features of crowds in complex scenes. 
This strategy provides the model with a unique perspective to observe and process image data, significantly reducing computational complexity while maintaining global perception capability. 
\begin{figure}[h]
	\centering
	\includegraphics[width=0.4\textwidth]{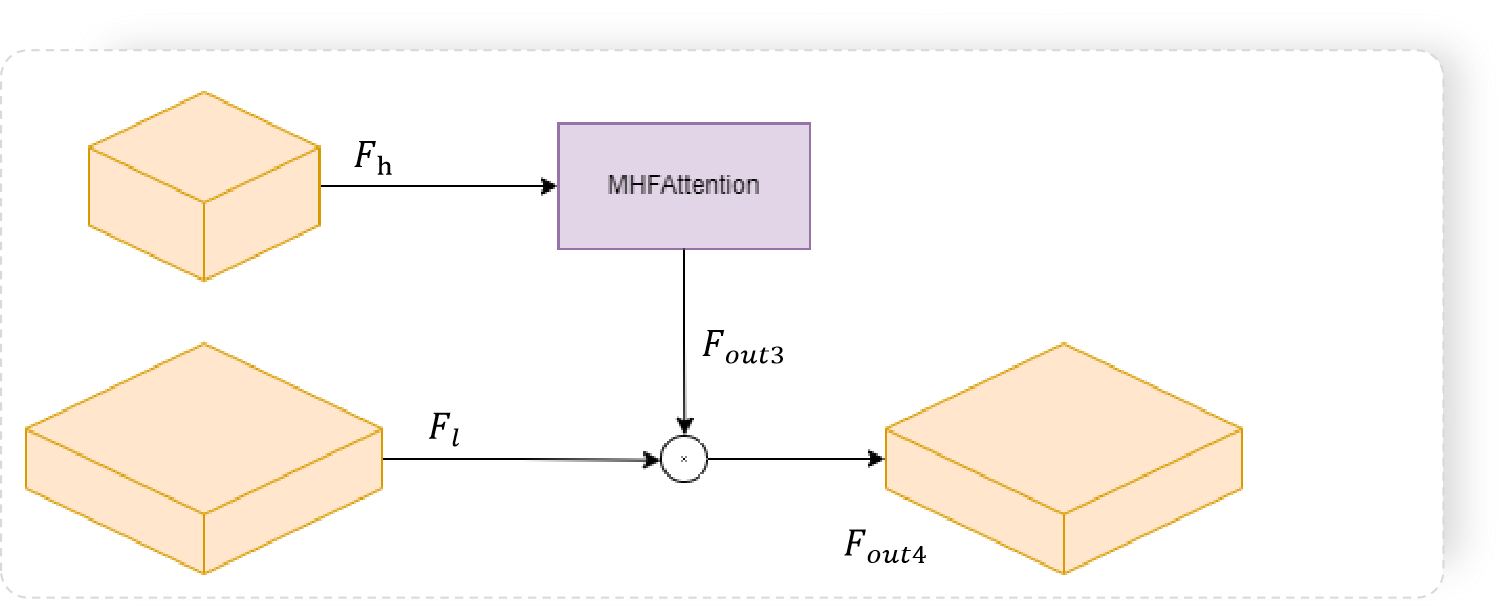}\hfill
	\caption{MHF Attention block connection.}
	\label{fig:03}
\end{figure}

MHF attention blocks enhance the model's ability to capture and represent important features by integrating high-level features when handling occlusion and multi-scale crowds.
Additionally, we also design the enhancement scheme for enhancing low-level features using high-level features, as shown in Fig. \ref{fig:03}. Let $\mathbf{F} _{l}$ be low-level features. Then, our enhancement scheme can be described as
\begin{equation}\label{eq:08}
\mathbf{F}_{out_4} = \mathbf{F}_l\odot\mathbf{F}_{out_3}
\end{equation}
where $\odot$ is the element-wise multiplication operator, $\mathbf{F}_{out_3}$ is the output of MHF attention block, which is obtained from high-level features $\mathbf{F}_h$. 
Ultimately, HS2FPN achieves the fusion of crucial feature extraction and representation capabilities for multi-scale features.

\begin{table*}[t]
    \centering
    \caption{Performance comparison of mainstream crowd counting models on three datasets.}
    \begin{tabular}{lcccccccccc}
        \toprule
        \multirow{2}{*}{Method} & \multirow{2}{*}{Localization} & \multirow{2}{*}{Architecture} & \multicolumn{2}{c}{SHT\_A} & \multicolumn{2}{c}{SHT\_B} & \multicolumn{2}{c}{UCF\_QNRF} & \multicolumn{2}{c}{JHU-Crowd} \\
        \cline{4-11}
        & & & MAE & MSE & MAE & MSE & MAE & MSE & MAE & MSE \\
        \midrule
        MCNN\cite{MCNN} & $\times$ & CNN & 110.2 & – & 26.4 & – & – & – & 160.6 & 377.7 \\
        CSRNet\cite{CSRNet} & $\times$ & CNN & 68.2 & 115.0 & 10.6 & 16 & 120.3 & 208.5 & 72.2 & 249.9 \\
        CANNet\cite{CANNet} & $\times$ & CNN & 62.3 & 100.0 & 7.8 & 12.2 & 107.0 & 183 & 100.1 & 314.0 \\
        ASNet\cite{ASNet} & $\times$ & CNN & 57.78 & 90.13 & – & – & 91.59 & 159.71 & – & – \\
        SDANet\cite{SDANet} & $\times$ & CNN & 63.6 & 101.8 & 7.8 & \uline{10.2} & – & – & \uline{59.3} & 248.9 \\
        SCARNet\cite{SCARNet} & $\times$ & CNN & 66.3 & 144.1 & 9.5 & 15.2 & – & – & – & – \\
        GauNet\cite{GauNet} & $\times$ & CNN & 54.8 & 89.1 & \textbf{6.2} & \textbf{9.9} & \textbf{81.6} & \textbf{153.7} & 69.4 & 262.4 \\
        P2PNet\cite{P2PNet} & $\surd$ & CNN & 52.7 & 85.1 & \uline{6.3}& \textbf{9.9} & 85.32 & 154.5 & – & – \\
        FGENet\cite{FGENet} & $\surd$ & CNN & \textbf{51.66} & 85.00 & 6.34 & 10.53 & \uline{85.2} & 158.76 & – & – \\\hline
        CLTR\cite{CLTR} & $\surd$ & Transformer & 56.9 & 95.2 & 6.5 & 10.6 & 85.8 & \uline{141.3} & 59.5 & \uline{240.6} \\
        LoViTCrowd\cite{LoViT} & $\times$ & Transformer & 54.8 & \textbf{80.9} & 8.6 & 13.8 & 87 & 141.9 & – & – \\ \hline
        VMambaCC & $\surd$ & Mamba & \uline{51.87} & \uline{81.3} & 7.48 & 12.47 & 88.42 & 144.73 & \textbf{54.41} & \textbf{201.93} \\
        \bottomrule
    \end{tabular}
    \label{tab:01_ccres}
\end{table*}
\section{Experiments}
To validate the effectiveness of VMambaCC, we performed experiments on four public crowd datasets and compared it with state-of-the-art models.  
\subsection{Experimental Setups}

\textbf{Datasets.} Four challenging datasets were used here: Shanghai Tech \cite{MCNN}, UCF\_QNRF \cite{QNRF}, JHU-Crowd \cite{JHU} and UCF\_CC\_50 \cite{UCFCC50}. The detailed information of these dataset is shown in Table \ref{tab:data}, where both SHT\_A and SHT\_B belong to the Shanghai Tech dataset.
\begin{itemize}
\item Shanghai Tech is a large-scale crowd counting dataset consisting of 1198 images and 330,165 annotations.
This dataset covers various scene types and densities. There is some imbalance in the number of images with different densities. SHT\_A and SHT\_B are its two subsets, where SHT\_A has significantly higher density than SHT\_B. 

\item UCF-QNRF is a dataset containing 1535 challenging images and 1,251,642 annotations, with diverse image scenes, angles, densities, and lighting variations.

\item JHU-Crowd is a large-scale dataset with 4250 images and 1,114,785 annotations, averaging 262 individuals per image. Additionally, this dataset provides rich annotations, including point-level, image-level, and head-level annotations.

\item UCF\_CC\_50 is a challenging dataset containing 50 images with an average of 1280 annotations per image. This dataset contains a variety of scenes and perspectives, such as concerts, protests, stadiums and marathons.
\end{itemize}

\textbf{Data Augmentation.} Some data augmentation strategies were adopted here, including random scaling with a probability of 0.5 (scaling factors ranging from 0.7 to 1.3), random horizontal flipping with a probability of 0.5, color adjustments, Gaussian noise addition, and random cropping to the size of $128\times 128$. For the QNRF dataset, we constrained the maximum size to not exceed 1408 and maintained the aspect ratio of the original images.

\textbf{Hyperparameter Settings.} The Adam \cite{adam} optimizer was applied to our model with a learning rate of $10^{-4}$. For each image in all datasets except QNRF, we set a reference point every two pixels. For the QNRF dataset, we generated a reference point every two pixels vertically and horizontally.

\textbf{Evaluation Metrics.} For crowd counting tasks, Mean Absolute Error (MAE) and Mean Squared Error (MSE) are two common evaluation metrics. MAE represents the average of the absolute errors between predicted and true values, while MSE is the average of the squared errors between predicted  and true values. These two metrics help us assess the accuracy and precision of the model in crowd counting tasks, which are defined as
\begin{align*}
    \text{MAE} = \frac{1}{NUM} \sum_{j=1}^{NUM} |N_j - M_j|
\end{align*}
and
\begin{align*}
    \text{MSE} = \frac{1}{NUM} \sqrt{\sum_{j=1}^{NUM} (N_j - M_j)^2}
\end{align*}
where $N_j$ and $M_j$ are the numbers of annotation and predicted points in the $j$-th test image, and $NUM$ is the number of test images. Lower values of the both metrics indicate better performance of a model.

For localization tasks, Precision (P), Recall (R), and F1-score (F1) are the evaluation metrics. 
Following \cite{CLTR}, we define $\sqrt{d_w^2+d_h^2}/2$ as the distance between annotated points and matched predicted points, where $d_w$ is the horizontal distance between the predicted and true points, and $d_h$ is the vertical distance between the matched predicted point and the annotated point. 
The annotated points are treated as positive samples. Now, we need to judge which matched predicted points are positive or negative. Let $\sigma$ be the judgment threshold.  
Specifically, a matched predicted point is positive if the distance between it and the corresponding annotated point is less than $\sigma$; otherwise, it is negative. In this case, these metrics can be described as
\[ P = \frac{TP}{N} \]
and
\[ R = \frac{TP}{M} \]
where $TP$ is the number of positive matched predicted points, $N$ is the number of annotated points, and $M$ is the number of predicted points. Because F1-score is the harmonic mean of precision and recall, it provides a single score that balances both of these measures, which is defined as 
\[ F1 = 2 \times \frac{P \times R}{P + R} \]

\begin{figure*}[t]
	\centering
	\includegraphics[width=0.9\textwidth]{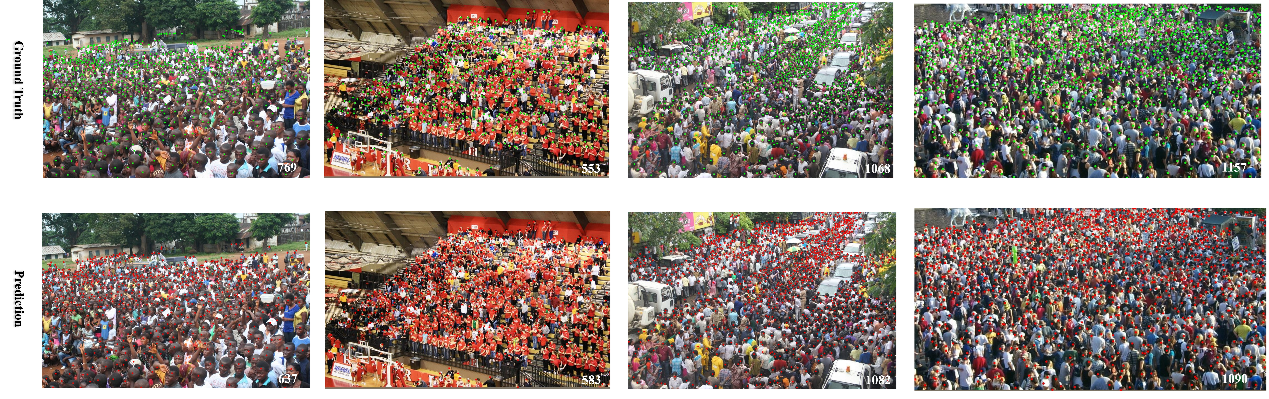}\hfill
	\caption{Demonstration of the predictive effect of our model.}
	\label{fig4:result}
\end{figure*}
\subsection{Counting Experiments}
For crowd counting tasks, VMambaCC was compared with 11 mainstream methods, including 9 CNN-based models (MCNN \cite{MCNN}, CSRNet \cite{CSRNet}, CANNet \cite{CANNet}, ASNet \cite{ASNet}, SDANet \cite{SDANet}, Spatial-Channel-wise Attention Regression (SCARNet) \cite{SCARNet}, Gaussian kernels Network (GauNet) \cite{GauNet}, P2PNet \cite{P2PNet}, and FGENet \cite{FGENet}) as well as 2 Transformer-based models (CLTR \cite{CLTR} and LoViTCrowd \cite{LoViT}).

Experimental results on Shanghai Tech, UCF\_QNRF, and JHU-Crowd datasets are shown in Table \ref{tab:01_ccres}, where the best values are in bold font, and the second best are underlined. From Table \ref{tab:01_ccres}, we can see that in terms of MSE, VMambaCC achieves the second best result on SHT\_A, which is 0.4 points lower compared to the best method LoVitCrowd. In terms of MAE, our method differs from FGENet by 0.21 points. We consider that there is some annotation inaccuracy in SHT\_A.

Additionally, GauNet (a CNN-based model) performs the best on SHT\_B and our model (a VMamba-based model) is not so good, which is possibly due to insufficient samples and fewer annotated points in SHT\_B. In this case, there has a certain gap between VMambaCC and existing methods.

On the UCF\_QNRF dataset, there is also a certain gap between VMambaCC and the current best method. Some samples in UCF\_QNRF have tens of thousands of annotations, which increases the computational effort of the Hungarian matching algorithm. Owing to experimental equipment limitations, our method cannot use an appropriate batch\_size in the training process; thus, VMambaCC could not achieve its best performance. 

VMambaCC achieves the best results on the JHU-Crowd dataset, with an improvement of 4.89 in MAE compared to the second-ranked SDANet method and an improvement of 38.67 in MSE compared to the second-ranked CLTR method.

On the UCF\_CC\_50 dataset, five-fold cross-validation was employed. Results are presented in Table \ref{tab:ucfcc50}. The comparison with other methods is listed in Table \ref{tab:cucfcc50}. Observation on Table \ref{tab:cucfcc50} indicates that our method achieves significant improvements over other approaches on the UCF\_CC\_50 dataset. Compared to the second-ranked FGENet, VMambaCC has an improvement of 78.22 points in MAE and 137.16 points in MSE. Obviously, Our method has reached the state-of-the-art level on this datasets. 

\begin{table}[htbp]
\centering
\caption{Five-fold cross-validation of VMambaCC on UCF\_CC\_50.}\label{tab:ucfcc50}
\begin{tabular}{ccc}
\hline
Fold  & MAE & MSE \\ \hline
1 & 84.00 & 103.96 \\ 
2 & 37.70 & 47.04 \\ 
3 & 53.90 & 66.84 \\ 
4 & 68.20 & 78.97 \\ 
5 & 77.90 & 96.75 \\
\hline 
Mean & 64.34 & 78.712 \\ \hline
\end{tabular}
\end{table}

\begin{table}[htbp]
\centering
\caption{Performance comparison of mainstream crowd counting models on UCF\_CC\_50.}\label{tab:cucfcc50}
\begin{tabular}{lcc}
\hline
Method & MAE & MSE \\ \hline
MCNN \cite{MCNN} & 377.6 & – \\ 
CSRNet \cite{CSRNet} & 266.1 & 397.5 \\ 
CANNet \cite{CANNet} & 212.2 & 243.7 \\ 
ASNet \cite{ASNet} & 174.84 & 251.63 \\ 
SDANet \cite{SDANet} & 227.6 & 316.4 \\ 
SCARNet \cite{SCARNet} & 227.6 & 316.4 \\
GauNet \cite{GauNet} & 186.3 & 256.5 \\
P2PNet \cite{P2PNet} & 172.7 & 256.1 \\ 
FGENet \cite{FGENet} & \uline{142.56} & \uline{215.87} \\ \hline
VMambaCC & \textbf{64.34} & \textbf{78.71} \\ \hline
\end{tabular}
\end{table}
\subsection{Localization Experiments}
To compare VMambaCC with other methods, we conducted localization experiments on SHT\_A because this dataset has a large scale variation. Comparison methods are Localization-based Fully CNN (LCFCN) \cite{LCFCN},Method in \cite{Methodin}, Locate-Size and Count CNN  (LSC-CNN) \cite{LSC-CNN}, Topological constraints Count (TopoCount) \cite{TopoCount}, and CLTR \cite{CLTR}. Here, we set $\sigma=4$ and  $\sigma=8$. 

Experimental results are shown in Table \ref{tab:shta}. We can see that VMambaCC achieves the best localization performance.
When $\sigma=4$, VMambaCC outperforms the second-best method by 16.2\% in precision, 17.6\% in recall, and 16.85\% in F1-score.
If $sigma=8$, our method surpasses the second-best method by 3.16\% in precision, 5.21\% in recall, and 4.18\% in F1-score. Findings demonstrate the effectiveness of VMamba in localization tasks.

To further validate the localization performance of VMambaCC, we also performed experiments on SHT\_B, JHU-Crowd, and UCF\_QNRF datasets. The results are shown in Table \ref{tab:loacl}. It is evident that our approach has achieved commendable performance across these three datasets. Moreover, the outcomes provide a benchmark for comparison that will be instrumental for subsequent research endeavors.

\begin{table}[htbp]
    \centering
     \caption{Localization performance comparison obtained by six methods on SHT\_A.}\label{tab:shta}
    \begin{adjustbox}{max width=1\linewidth,width=0.95\linewidth}
    \begin{tabular}{lccc|ccc}
        \hline
        \multirow{2}{*}{Method} & \multicolumn{3}{c}{$\sigma$=4} & \multicolumn{3}{|c}{$\sigma$=8} \\ \cline{2-4} \cline{5-7}
         & P(\%) & R(\%) & F(\%) & P(\%) & R(\%) & F1(\%) \\ \hline
        LCFCN \cite{LCFCN} & 43.30\% & 26.00\% & 32.50\% & 75.10\% & 45.10\% & 56.30\% \\ 
        Method in \cite{Methodin} & 34.90\% & 20.70\% & 25.90\% & 67.70\% & 44.80\% & 53.90\% \\ 
        LSC-CNN \cite{LSC-CNN} & 33.40\% & 31.90\% & 32.60\% & 63.90\% & 61.00\% & 62.40\% \\ 
        TopoCount\cite{TopoCount} & 41.70\% & 40.60\% & 41.10\% & 74.60\% & 72.70\% & 73.60\% \\ 
        CLTR \cite{CLTR} & \uline{43.60\%} & \uline{42.70\%} & \uline{43.20\%} & \uline{74.90\%} & \uline{73.50\%} & \uline{74.20\%} \\ \hline
        VMamba  & \textbf{59.80\%} &\textbf{ 60.3\%} & \textbf{60.05}\% & \textbf{78.06\%} &	\textbf{78.71\% }&	\textbf{78.38\%} \\ \hline
    \end{tabular}
   \end{adjustbox}
\end{table}

\begin{table}[htbp]
\centering
\caption{Localization experiments of VMambaCC on different datasets.}\label{tab:loacl}
\begin{tabular}{lcccc}
\hline
Dataset & $\sigma$ & P(\%) & R(\%) & F(\%) \\ \hline
SHT\_B & 4 & 63.77\% & 63.21\% & 63.49\% \\ 
SHT\_B & 8 & 80.77\% & 80.06\% & 80.41\% \\ 
JHU-Crowd & 4 & 47.89\% & 46.18\% & 47.02\% \\ 
JHU-Crowd & 8 & 64.63\% & 62.32\% & 63.45\% \\ 
UCF\_QNRF & 4 & 43.59\% & 42.91\% & 43.25\% \\ 
UCF\_QNRF & 8 & 69.75\% & 68.67\% & 69.21\% \\ \hline
\end{tabular}
\end{table}

\subsection{Ablation experiments}

\textbf{MHF.} Simple ablation experiments about the MHF attention mechanism were conducted on the SHT\_A dataset . The possible ablation components are CEM, MSEM, HLCEM, the number of heads, and connection position. Thus, we ablated them separately, as shown in Table \ref{tab:MHHLF}. 

From Table \ref{tab:MHHLF}, it can be seen that the baseline model without MHF (the first row in this table) obtained an MAE of 54.75 and an MSE of 88.46 on the SHT\_A dataset. When adding CEM alone (the second row in Table \ref{tab:MHHLF}), the model performance deteriorates compared to the baseline because some noise may be introduced through directly multiplying channels of high-level features with low-level features. 
However, when adding both CEM and MSEM with one head (the third row in Table \ref{tab:MHHLF}), both MAE and MSE of this model are improved  because this operation retains spatial information in high-level features.
Additionally, setting Head Nums to 4, the fusion effect of multiple sets of weights extracted by MSEM is not good, resulting in worse results compared to no multi-head operation. 
To determine the correct position for element-wise multiplication, we compared feature enhancement before and after feature fusion. The results show that performing element-wise multiplication after feature fusion leads to crossover effects, which may introduce noise. Thus, it is better to perform the operation of element-wise multiplication before feature fusion.

\begin{table}[h]
\centering
\caption{Ablation Experiments on MHF}\label{tab:MHHLF}
\begin{adjustbox}{max width=1\linewidth,width=0.9\linewidth}
\begin{tabular}{c c c c c c c }
\hline
\multirow{2}{*}{CEM} & \multirow{2}{*}{MSEM} & \multirow{2}{*}{HLCEM} & \multirow{2}{*}{\#Head} & \multirow{2}{*}{Connection position} & \multicolumn{2}{c}{SHT\_A} \\
\cline{6-7}
& & & & & MAE & MSE \\ \hline
$\times$ & $\times$ & $\times$ & -- & -- & 54.75 & 88.46 \\ 
$\checkmark$ & $\times$ & $\times$ & -- & before & 61.71 & 101.31 \\ 
$\checkmark$ & $\checkmark$ & $\times$ & 1 & before & 52.16 & 83.92 \\ 
$\checkmark$ & $\checkmark$ & $\times$ & 1 & post & 52.21 & 87.85 \\ 
$\checkmark$ & $\checkmark$ & $\times$ & 4 & before & 54.3 & 90.83 \\ 
$\checkmark$ & $\checkmark$ & $\checkmark$ & 4 & post & 63.88 & 97.52 \\ 
$\checkmark$ & $\checkmark$ & $\checkmark$ & 4 & before & 51.87 & 81.30 \\ \hline
\end{tabular}
\end{adjustbox}
\end{table}

\begin{table}[htbp]
\centering
\caption{Ablation experiments on stage of feature fusion.}\label{tab:fpn}
\begin{tabular}{lcc}
\hline
\multirow{2}{*}{Feature fusion} & \multicolumn{2}{c}{SHT\_A}   \\ 
 & MAE & MSE \\ \hline
FPN \cite{FPN} & 60.53 & 105.36 \\ 
FGFP \cite{FGENet} & 56.28 & 90.83 \\ \hline
HS2FPN (Without MHF)& 54.75 & 88.46 \\
HS2FPN (ours) & \textbf{51.87} & \textbf{81.30} \\ \hline
\end{tabular}
\end{table}

\textbf{Feature Fusion}
Comparative experiments on the feature fusion part were also conducted. Our feature fusion network could be replaced by other networks, such as FPN \cite{FPN} and FGFP \cite{FGENet}. In addition, our HS2FPN could delete the MHF attention mechanism. Thus, four feature fusion networks were compared, as shown in Table \ref{tab:fpn}. 
 
When using FPN in the feature fusion stage, our model  obtains 60.53 MAE and 105.36 MSE. We believe that FPN simply adds different features may lead to the loss of many important semantic information.
VMambaCC with FGFP obtains 56.28 MAE and 90.83 MSE, which is better than VMambaCC with FPN. The improvement reason lies in the fact that FGFP weights the spatial and channel dimensions, allowing the model to focus more on important information. 
VMambaCC with HS2FPN has the best results in both MAE and MSE. Compared to FGFP, HS2FPN shows improvements of 4.41 MAE and 9.53 MSE. This advantage is mainly due to our reasonable utilization of high-level semantic features to supervise low-level semantic features.

\subsection{Visualization}
The comprehensive set of experiments has substantiated the efficacy of our approach in the domain of counting and localization. These validations not only affirm the dependability of our method but also lay down a solid theoretical groundwork for the advancement and refinement of future models. The results of our predictions are shown in Figure \ref{fig4:result}. We are confident that these achievements will significantly propel research in the field and lay the foundation for subsequent technological innovations.
\section{Conclusion}
In this study, we introduce VMamba and propose an innovative crowd counting model, VMambaCC. This counting model has a global receptive field and the linearly escalating computational complexity in the context of crowd counting. 
A new attention mechanism, MHF, is design in this study. 
MHF employs high-level semantics to guide the refinement of lower-level semantic details; thus, it can avoid the loss of high-level semantic information during feature fusion.
By leveraging the MHF mechanism as a foundation, we have further developed the feature fusion network, HS2FPN.  
Extensive experimental validation confirms the pronounced superiority of our model in confronting contemporary challenges in crowd counting and localization. 
Therefore, VMambaCC contributes significant insights and a reference framework for further research and practical applications within crowd counting domain.


\newpage
\bibliographystyle{ACM-Reference-Format}
\bibliography{sample-base}

\end{document}